\documentclass[runningheads]{llncs}
\usepackage[T1]{fontenc}
\usepackage{cite}
\usepackage{amsmath} 
\usepackage{hyperref}
\usepackage[a4paper, margin=1.4in]{geometry}
\usepackage{booktabs}
\usepackage{caption}
\usepackage{graphicx} 
\usepackage{subcaption} 

\usepackage[T1]{fontenc}
\usepackage{graphicx}
\usepackage{color}

\begin{document}
\title{VS-DDPM: Efficient Low-Cost Diffusion Model for Medical Modality Translation}
%
%

\author{Nikoo Moradi \inst{2} \and
Gijs Luijten \inst{2,5,9} \and
Behrus Hinrichs-Puladi \orcidID{0000-0001-5909-6105} \inst{6,7} \and
Jens Kleesiek\inst{2,3,4,8,10} \and Victor Alves\orcidID{0000-0003-1819-7051}\inst{1} \and  
Jan Egger\inst{2,3,5,9,10} \and 
André Ferreira \inst{1,2,6,7}\orcidID{0000-0002-9332-0091}* 
}
\authorrunning{N. Moradi et al.}
%
\institute{Center Algoritmi / LASI, University of Minho, Braga, 4710-057,  Portugal
\and
Institute for Artificial Intelligence in Medicine (IKIM), Essen University Hospital
(AöR), University of Duisburg-Essen, Essen, Germany
\and 
Cancer Research Center Cologne Essen (CCCE), West German Cancer Center,
University Hospital Essen (AöR), Essen, Germany
\and 
German Cancer Consortium (DKTK), Partner site University Hospital Essen
(AöR), Essen, Germany
\and 
Institute of Computer Graphics and Vision, Graz University of Technology, Inffeldgasse 16, Graz, 8010, Austria
\and 
Institute of Medical Informatics, University Hospital RWTH Aachen, Aachen, Germany\\
\and 
Department of Oral and Maxillofacial Surgery, University Hospital RWTH Aachen, Aachen,  Germany \\
\and 
Department of Physics, TU Dortmund University, Dortmund, Germany \\
\and
Center for Virtual and Extended Reality in Medicine (ZvRM), University Hospital Essen (AöR), Essen, Germany \\
\and 
Faculty of Computer Science, University of Duisburg-Essen, Essen, Germany
\email{*andrefilipe.desousaferreira@gmail.com}}

\maketitle              
\begin{abstract}
Diffusion models produce high-quality synthetic data but suffer from slow inference. We propose 3D Variable-Step  Denoising Diffusion Probabilistic Model (VS-DDPM) a framework engineered to maintain generative quality while accelerating inference by several factors.
We tested our approach on four tasks—missing MRI, tumor removal, MRI-to-sCT, and CBCT-to-sCT—within the BraTS2025 and SynthRAD2025 challenges. Designed for high efficiency under hardware and time constrains imposed by both challenges.
VS-DDPM achieved state-of-the-art (SOTA) performance in missing MRI synthesis, yielding Dice scores of 0.80, 0.83, and 0.88 for the enhancing tumor, tumor core, and whole tumor regions, respectively, alongside a structural similarity index (SSIM) of 0.95. For MRI tumor removal, the model attained a root mean squared error (RMSE) of 0.053, a peak signal-to-noise ratio (PSNR) of 26.77, and an SSIM of 0.918. While the framework demonstrated competitive performance in MRI-to-sCT and CBCT-to-sCT tasks, it did not reach SOTA benchmarks, potentially due to sensitivities in data pre and post-processing pipelines or specific loss function configurations. These results demonstrate that VS-DDPM provides a robust and tunable solution for high-fidelity 3D medical image synthesis. The code is available in \href{https://github.com/andre-fs-ferreira/SynthRAD_by_Faking_it}{GitHub}.

\keywords{BraTS 2025 \and SynthRAD2025  \and Diffusion Models \and Medical Imaging \and Translation \and Inpainting}
\end{abstract}
\section{Introduction}
Medical imaging plays a central role in the diagnosis and treatment of cancer patients, particularly in radiotherapy (RT), which is used in more than half of all cancer cases \cite{Spadea2021Deep}. In RT, pre-treatment imaging is performed to develop a treatment plan, which is then delivered in daily fractions over several weeks. Traditionally, X-ray-based imaging, such as computed tomography (CT), has been the primary modality for patient positioning and monitoring before, during, or after treatment \cite{Chernak1975}. CT provides accurate anatomical geometry and essential electron density information for dose calculation. However, repeated CT scans expose patients to ionizing radiation, offer limited soft-tissue contrast, and are often not available directly in treatment rooms \cite{Edmund2017MRIonly}. To overcome these limitations, other imaging methods are increasingly being used.

Magnetic resonance imaging (MRI) offers superb soft-tissue contrast without ionizing radiation \cite{Schmidt2015MRI}. However, MRI scans lack the necessary attenuation data for dose calculations and optimization of the treatment plan \cite{Boldrini2017}. Cone-beam computed tomography (CBCT) is widely used for in-room positioning and monitoring \cite{BodaHeggemann2011}, yet suffers from shading, streaking, and other artifacts that reduce image quality and limit its suitability for dose calculation \cite{Ramella2017}. Converting MRI or CBCT to synthetic CT (sCT) addresses these limitations by enabling accurate dose calculations from modalities with superior soft-tissue visualization or convenient in-room acquisition \cite{Spadea2021Deep}.

Efforts such as the  SynthRAD2025 Grand Challenge \cite{Thummerer2025SynthRAD2025}\footnote{\href{https://synthrad2025.grand-challenge.org/synthrad2025/}{https://synthrad2025.grand-challenge.org/synthrad2025/}}, provide large datasets focusing on finding solutions for reducing the amount of radiation, by converting CBCT into CT or removing radiation entirely with MRI to CT conversion. 

Even when MRI only can be used for treatment planning, e.g. brain glioma, the long-time sessions and patient related constrains limits the quality of some modalities acquired, as 4 MRI modalities are necessary for a full informed decision. Therefore, generating the missing/low quality modality would increase the speed of diagnosis and treatment planing.

The existence of morphological distortions, e.g. brain gliomas, also affects other treatments. This is particularly important for algorithms that have only been trained on healthy brains, such as brain extraction, tissue segmentation and brain anatomy parcellation algorithms. It is also important to monitor the expected progression of the brain tumor during treatment and to study biomarkers by comparing the healthy and unhealthy brain in other diseases such as Alzheimer’s. 

The BraTS2025 \footnote{\href{https://www.synapse.org/Synapse:syn64153130}{https://www.synapse.org/Synapse:syn64153130}} provides a platform and several datasets aimed to help build solutions to solve these issues. With the tools provided by SynthRAD2025 and BraTS2025 datasets is possible to build solutions for:

\textbf{MRI-only RT:} sCTs derived from MRI enable treatment planning and delivery without the need for a separate CT scan, thereby eliminating registration errors between MRI and CT, reducing radiation exposure, and simplifying clinical workflows \cite{Edmund2017MRIonly}.

\textbf{Adaptive MRI- or CBCT-based RT:} sCTs generated from MRI or CBCT allow accurate dose calculations on daily imaging, enabling treatment plan adaptation to anatomical changes throughout the course of therapy \cite{Spadea2021Deep}.

\textbf{Faster MRI acquisition:} Missing MRI modalities can be generated allowing algorithms that expect all modalities work correctly \cite{hu2022domain}.

\textbf{Brain correction:} Removal of brain lesions that might affect treatment planing of other conditions \cite{kofler2023brain}.

To solve these challenges, we propose Variable-Step Denoising Diffusion Probabilistic Models (VS-DDPM), a high-quality and low resource generation pipeline that ensure realism and utility. Our approach is tested in 4 tasks: MRI to sCT, CBCT to sCT, missing modality generation and tumor removal. The code is available in \href{https://github.com/andre-fs-ferreira/SynthRAD_by_Faking_it}{GitHub}\footnote{\href{https://github.com/andre-fs-ferreira/SynthRAD_by_Faking_it}{https://github.com/andre-fs-ferreira/SynthRAD\_by\_Faking\_it}}.

\section{Material and Methods}

\subsection{Dataset}
For the MRI to sCT and CBCT to sCT we use the datasets provided by the SynthRAD2025 Grand Challenge, a multi-center dataset designed to benchmark sCT generation methods\cite{synthRAD2025dataset}.  The full dataset comprises 890 MRI-CT pairs, and 1,472 CBCT-CT pairs sourced from five European university medical centers. The data covers three anatomical regions: Head and Neck (HN), Thoracic (TH), and Abdominal (AB). The dataset is divided into training (65\%), validation (10\%), and testing (25\%) subsets. For our research, we only used the publicly available training data, as the validation and testing ground truth images were withheld by the challenge organizers.

For the MRI missing modality generation, the dataset is composed of 1251 training cases from the BraTS 2023 Glioma dataset \cite{baid2021rsna} and 238 training cases from the BraTS 2023 MET dataset \cite{moawad2024brain}. The modalities T1, T1-Gd, T2 and FLAIR and the respective segmentations are available for each case. For the tumor removal, the BraTS 2023 Glioma \cite{baid2021rsna} dataset is also used, but only the T1 modality. No additional public datasets were used in this research.

\subsubsection{Data Preprocessing:}

The SynthRAD2025 datasets are provided with rigid registration to align the MRI/CBCT scans with the corresponding ground truth CT images, and defaced for anonymity. We performed deformable registration and implemented data augmentation to assess performance improvements. All modalities (CT, MRI, and CBCT) underwent spatial transformations, including 3 degree rotations, scaling, and grid distortion. MRI data was further augmented with intensity-specific perturbations: bias field effects, intensity shifts, and Gaussian smoothing/noise.

For the BraTS 2023 dataset (already skull-stripped and rigidly registered), we employed the MRI augmentation used in the SynthRAD experiments, with the inclusion of shearing. To build the ground truth of the tumor removing task, random portions of the healthy region of the brain were selected using real tumor masks, creating a very large and diverse dataset.

\subsubsection{Data Splitting:}
We use the training dataset available and created a custom 90\%/10\% train-validation split using a uniformly random selection process. This ensured that our validation set included a representative number of cases from all anatomical regions and all participating medical centers.

\subsubsection{Data Normalization:} 
For CT and CBCT normalization, we clip the intensities from a range of -1000 HU to 1600 HU and normalized linearly to a range of -1 to 1. 
The MRI scans were normalized using percentile clipping of 0.1\% and 99.9\%, z-scoring and linear normalization between -1 and 1. For the missing modality task z-scoring was applied individually to each modality with global statistics. For the tumor removing task, the statistics for z-scoring where computed using the individual scan without the tumor and considering only non-zero values. For the MRI to sCT translation task the statistic were computed individually per case.
Using the z-score reduces the impact of MRI scans from multiple facilities, resulting in greater stability than normal linear normalization. Z-scoring of the CT and CBCT scans was also tested, but produced poorer results.

\subsection{Models and Training processes}
\subsubsection{Denoising Diffusion Probabilistic Models:}

Denoising Diffusion Probabilistic Models (DDPMs) have shown great potential for generating new cases. DDPMs have achieved even better results than GANs in generating natural images \cite{dhariwal2021diffusion}.
However, DDPMs rely on the Markov chain to achieve good results, making the process more computationally intensive and slower. This is aggravated when it comes to volumetric medical images. Therefore, we have developed our solution based on the Improved DDPMs (IDDPMs) \cite{nichol2021improved}. 

To make our solution less computational demanding, we optimize it to run in less than 15 minutes, as expected by the SynthRAD challenge. Due to the variable size of each scan in the MRI/CBCT to sCT translation, our method is optimized to work with variable number of sampling step, making each inference in less than 15 minutes. Therefore, we developed a Variable-Step approach (VS-DDPM) that trains the model to support multiple sampling-step settings, making inference both more robust and faster without further fine-tuning, and allowing longer $T$ on faster GPUs (or smaller scans) and shorter $T$ on slower GPUs (or larger scans). This allows for faster training as the model is specialized in the steps needed, and faster generation that matches the requirements of the end machine, in this case the NVIDIA T4 GPUs available in the challenges platforms. This is not an issue with the BraTS based dataset, as all scans have the same shape. 

The 3D Swin-ViT \cite{pan2024synthetic} and a 3D U-Net \cite{ronneberger2015u} were tested. Due to the variable size of the input, using the entire volume was not feasible, especially for the heavier networks such as the 3D Swin-ViT. Therefore, we opted for patched-based training and a sliding-window approach for inference. Cosine learning rate scheduling and early stopping were used in the training process. 

For the SynthRAD datasets, the training of the VS-DDPM uses random number of $T$ steps, selected and resampled at each epoch. Due to the high variability of the dataset, the values 5, 10, 15, 20, 25, 35, 50, 75, 100, 125, 150, 175, 200, 225, 250, 275 and 300 were used. 
For the BraTS datasets, fixed 25 steps were used. More steps were tested, but without improvements of the results .

The base loss function is composed by $L_{MAE}$ and $\lambda L_{vlb}$  for all the tasks \cite{nichol2021improved}. $L_{MSE}$ and $L_{SSIM}$ are also used in the BraTS related tasks. The use of Dice Score (DSC) loss for the SynthRAD related challenges  was also tested, but since it did not improve the results and introduced instability this loss was abandoned. Instead, we use the Anatomical Feature-Prioritized (AFP) loss \cite{longuefosse2024anatomical} for fine-tuning the model for the MRI/CBCT to sCT task. 

Given the predicted volume 
$\tilde{x}_0$ and the ground-truth volume $x_0$, the total losses are defined by Equations \ref{eq:brats} and \ref{eq:synthrad} for the BraTS and SynthRAD related tasks, respectively.

\begin{equation}
L_{BraTS2025} =
\,L_{MAE}(\tilde{x}_0, x_0) +
\,L_{MSE}(\tilde{x}_0, x_0) + \,(1 - L_{SSIM}(\tilde{x}_0, x_0)) +
\lambda_1\,L_{\mathrm{VLB}}(\tilde{x}_0, x_0)
\label{eq:brats}
\end{equation}

\begin{equation}
L_{SynthRAD} =
\,L_{MAE}(\tilde{x}_0, x_0) +
\,\lambda_2L_{AFP}(\tilde{x}_0, x_0) +
\lambda_1\,L_{\mathrm{VLB}}(\tilde{x}_0, x_0)
\label{eq:synthrad}
\end{equation}

Where $\mathrm{MSE}$ and $\mathrm{MAE}$ denote the Mean Squared and Mean Absolute Errors, 
respectively; $\mathrm{SSIM}$ is the Structural Similarity Index Measure
(expressed here as a loss term via $1-\mathrm{SSIM}$); and $L_{\mathrm{VLB}}$ is the 
Variational Lower Bound term used in diffusion models to ensure probabilistic  consistency across denoising steps. We set $\lambda_1 = 0.001$  to regularize the diffusion process without dominating the reconstruction loss. 
The BraTS related tasks are trained fully in one complete phase. For the SynthRAD tasks, a two phase approach is used: First only MAE is used with a small (0.0001 weight) squared loss of the variation predicted by the model, to avoid too large and instable variation predictions; For fine-tunning, the MAE is kept and the $AFP$ loss added, with weights 1 and 0.2 respectively. The data augmentation is deactivated in this phase.

For the interference, we compute how much time it takes for each step to be completed on the available GPU, compute how many windows are needed to generate a complete volume and interpolate the maximum possible T, using Equation \ref{eq:find_T}. Then, we select the largest T value from the training steps that does not exceed the calculated T and re-calculate the window overlap possible for this final T.

\begin{equation}
T=\frac{Total\_time}{Time\_per\_infer*number\_windows}
\label{eq:find_T}
\end{equation}

Where total time is 900 seconds, time per infer is approximately 0.433 seconds in the NVIDIA T4 GPU and number of windows is defined by Equation \ref{eq:n_windows}.

\begin{equation}
\text{Number of windows} \;=\; \prod_{i=1}^3 \left(
\left\lceil \frac{I_i - R_i}{R_i (1 - p)} \right\rceil + 1
\right)
\label{eq:n_windows}
\end{equation}

Where
 \(I_1, I_2, I_3\) are the input sizes in each dimension,
 \(R_1, R_2, R_3\) are the window sizes, and \(p\) is the overlap fraction (\(0 \le p < 1\)).

Training continued until the first signs of overfitting and the validation metrics stabilized. For the MRI-CT task, the final model was trained for 3000 epochs with early stopping active after 2000 epochs, and 1400 epochs for the missing MRI modality and tumor removal. The reduced number of epochs for the last two tasks is justifiably by the lower complexity of the tasks, achieving convergence faster. AdamW with learning rate of 2e-5, betas 0.9, 0.999, and weight decay of 1e-5 with cosine learning rate scheduling down to 1e-6 was used for all the tasks.

\subsection{Evaluation}

The MRI/CBCT to sCT task was evaluated using \textit{image similarity} (MAE, PSNR and MS-SSIM) and \textit{geometric consistency} (DSC and HD95) as detailed in the SynthRAD repository \footnote{\href{https://github.com/SynthRAD2025/metrics}{https://github.com/SynthRAD2025/metrics}}. The missing modality task was evaluated with DSC and SSIM, and the tumor removing task with MSE, PSNR and SSIM.

\section{Results and Discussion}

\subsection{MRI/CBCT to sCT translation}

Table \ref{tab:synthrad_results_local} compares the performance of the networks across fine-tuning, step strategies, and overlap configurations. Although fine-tuning with AFP loss was expected to improve Dice scores and HD95, no significant improvement was observed. Nevertheless, the use of a variable number of steps yields results comparable to those of fixed sampling steps. This demonstrates the potential for adapting VS-DDPM to various input sizes and computational constraints.

\begin{table}[ht]
\footnotesize
\centering
\caption{Local validation results of MRI to sCT task. Our proposed method is presented in the last row. \textbf{Bold} = best.}
\begin{tabular}{@{}ccccccccc@{}}
\toprule
\textbf{Method} & \textbf{Network} & \textbf{Finetune} & \textbf{Step/Overlap} & \textbf{MAE} & \textbf{PSNR} & \textbf{MS-SSIM} & \textbf{Dice} & \textbf{HD95} \\
\midrule
 VS-DDPM &   \\
\midrule

DDPM & Swin-ViT  & No & 25/0.5 & 76.183 & 28.104 & 0.9076 & 0.7021 & 7.690  \\%
DDPM & Swin-ViT  & Yes & 25/0.5 & 73.805 & 28.301 & 0.9116 & 0.7071 & 7.279  \\%

VS-DDPM & Swin-ViT  & No & Variable &  77.220 & 28.052 & 0.9072 & 0.6988 & 7.530 \\%
VS-DDPM & Swin-ViT  & Yes & Variable & 74.400 & 28.270 & 0.9111 & 0.7072 & \textbf{7.199} \\%

DDPM & U-Net & No & 25/0.5 & 73.122 & 28.364 & 0.9137 & \textbf{0.7147} & 7.356 \\%
DDPM & U-Net & Yes & 25/0.5 & \textbf{71.969} & \textbf{28.508} & \textbf{0.9150} & 0.7112 & 7.330 \\%

VS-DDPM & U-Net & No & Variable & 73.895 & 28.322 & 0.9131 & \underline{0.7125} & 7.244 \\%
VS-DDPM & U-Net & Yes & Variable & \underline{72.499} & \underline{28.477} & \underline{0.9148} & 0.7106 & \underline{7.243} \\%
\bottomrule
\end{tabular}
\label{tab:synthrad_results_local}
\end{table}

Table \ref{tab:synthrad_results_online_test} presents a comparison between our method (same as the last row of Table \ref{tab:synthrad_results_local}) and the best method on the online testing platform (results available in the \href{https://synthrad2025.grand-challenge.org/evaluation/post-challenge-task-1-mri/leaderboard/}{SynthRAD2025 challenge platform}). All models take into account that only one NVIDIA T4 GPU is available and that the model must run for less than 15 minutes per case. The use of U-Net achieved the best results as expected, however, the results were consistently worse than the solution with best performance on the online platform.

\begin{table}[ht]
\footnotesize
\centering
\caption{\href{https://synthrad2025.grand-challenge.org/evaluation/post-challenge-task-1-mri/leaderboard/}{Online testing results} of MRI to sCT translation. Our proposed method is presented in the last row, with fine-tuning, variable step and overlap.}
\begin{tabular}{@{}ccccccccc@{}}
\toprule
\textbf{Method} & \textbf{Network} & \textbf{MAE} & \textbf{PSNR} & \textbf{MS-SSIM} & \textbf{Dice} & \textbf{HD95} \\
\midrule
Winner & resUnet-MAP-ensemble & 65.081 & 29.811 & 0.9345 & 0.7810 & 5.694 \\
VS-DDPM & U-Net & 68.732 & 29.072 & 0.9267 & 0.7583 & 6.883 \\
\toprule
\end{tabular}
\label{tab:synthrad_results_online_test}
\end{table}

DDIM was also tested, but the results are consistently worse, therefore, we did not include it in the table. To speed up inference, less overlap was also tested, but the results were worst due to the creation of artifacts at the edges of each window. Therefore, we ensured that the overlap was at lest 0.5. The cosine noise scheduler was also tested, but its use during the training of the VS-DDPM was very unstable, so the results are not shown.

For CBCT to sCT translation we chose the same pipeline and same parameters. Table \ref{tab:val_task2} presents the results on the local validation set, and Table \ref{tab:test_task2} the test results from the online test set (results available in the \href{https://synthrad2025.grand-challenge.org/evaluation/post-challenge-task-2-cbct/leaderboard/}{SynthRAD2025 challenge platform}).

From the results of Tables \ref{tab:synthrad_results_local} and \ref{tab:val_task2} can be seen that the use of variable steps leads to competitive results compared to the use of fixed 25 steps. This is particularly evident in the CBCT to sCT task, where variable steps achieved superior overall results. We noticed that the use of bigger window sizes improved consistently the Dice and HD95 metrics without sacrificing the MAE metric, however, it increases the computational footprint. Therefore, we decided to use 192x192x32 for MRI to sCT and 128x128x32 for the CBCT to sCT. We also concluded that more complex networks produce better results, even though less timesteps can be used. We carefully tuned the training process as well as the network to avoid overfitting, although some overfitting was seen. 

Since high efficiency is required for both training and inference, mixed precision was used, which increases the instability of the training. The same was observed when the cosine noise scheduler was used. We believe that using full precision would give better results, however, it would increase the training and inference time. We also found that the model trained for 25 steps gives better results than the model trained for 1000 steps. This can be explained by the faster convergence when training a model to predict only 25 steps compared to a model to predict 1000 steps, as the same number of epochs were used.

\begin{table}[ht]
\footnotesize
\centering
\caption{Local validation results on the CBCT to sCT task. Our proposed method is presented in the last row. \textbf{Bold} = best.}
\begin{tabular}{@{}ccccccccc@{}}
\toprule
\textbf{Method} & \textbf{Network} & \textbf{Finetune} & \textbf{Step/Overlap} & \textbf{MAE} & \textbf{PSNR} & \textbf{MS-SSIM} & \textbf{Dice} & \textbf{HD95} \\
\midrule
 VS-DDPM &   \\
\midrule

DDPM & Swin-ViT  & No & 25/0.5 & 299.785 & 21.1343 & 0.7475 & 0.2114 & 36.932  \\%
DDPM & Swin-ViT  & Yes & 25/0.5 & 59.096 & 29.850 & 0.9522 & 0.7826 & 3.698  \\%

VS-DDPM & Swin-ViT  & No & Variable &  307.099 & 21.038 & 0.7459 & 0.2105 & 36.031 \\%
VS-DDPM & Swin-ViT  & Yes & Variable & 59.067 & 29.860 & 0.9524 & 0.7844 & \textbf{3.570} \\%

DDPM & U-Net & No & 25/0.5 & 60.821 & 29.777 & 0.9515 & 0.7816 & \underline{3.670} \\%
DDPM & U-Net & Yes & 25/0.5 & \underline{58.415} & \underline{29.9175} & \underline{0.9529} & \textbf{0.7883} & 3.7499 \\%

VS-DDPM & U-Net & No & Variable & 60.930 & 29.779 & 0.9515 & 0.7819 & 3.676 \\%
VS-DDPM & U-Net & Yes & Variable & \textbf{58.289} & \textbf{29.939} & \textbf{0.9532} & \underline{0.7853} & 3.687 \\%
\bottomrule
\end{tabular}
\label{tab:val_task2}
\end{table}

\begin{table}[ht]
\footnotesize
\centering
\caption{\href{https://synthrad2025.grand-challenge.org/evaluation/post-challenge-task-2-cbct/leaderboard/}{Online testing results} of CBCT to sCT translation. Our proposed method is presented in the last row, with fine-tuning, variable step and overlap.}
\begin{tabular}{@{}ccccccccc@{}}
\toprule
\textbf{Method} & \textbf{Network} & \textbf{MAE} & \textbf{PSNR} & \textbf{MS-SSIM} & \textbf{Dice} & \textbf{HD95} \\
\midrule
Winner & resUnet-MAP-ensemble & 47.182 & 32.938 & 0.9690 & 0.8580 & 4.543 \\
VS-DDPM & U-Net & 50.165 & 32.075 & 0.9650 & 0.8460 & 4.981 \\
\toprule
\end{tabular}
\label{tab:test_task2}
\end{table}

\subsection{Missing modality generation and tumor inpainting}
For the missing modality generation the online validation was not available, therefore we only present the local validation and the online testing results.
Table \ref{tab:modal_generation_local} presents the results on the local validation set and the comparison of the the DSC with the use of all data for inference. Evaluation of the missing modality generator via downstream segmentation revealed a performance retention of over 95\%, with only a 4.5\% reduction in DSC compared to the full-modality baseline. The slight degradation indicates a minor loss of information during the generation process, a finding further validated by the SSIM analysis. Nevertheless, the proximity of these results to the baseline suggests that the synthesized data provides a robust representation of the missing structural information. Post-processing with clipping values bellow 0.01 (when normalized between 0 and 1) was necessary for this task, as the small background noise generated was degrading the performance of the segmentation model.

Table \ref{tab:modal_generation_online_test} presents the testing results with the standard deviation. Our approach achieves a new State-Of-The-Art (SOTA). We cannot share the other contestant results, only the \href{https://www.synapse.org/Synapse:syn64153130/wiki/633062}{rankings}. In contrast to the local validation set, the generated and real modalities exhibit high structural similarity in the test set, as evidenced by the high SSIM metrics.

\begin{table}[h!]
\centering
\caption{Performance comparison of missing modality generation on the local validation set using DSC and SSIM metrics.}
\begin{tabular}{llcccc}
\hline
\textbf{Name} & \textbf{Network} & \textbf{Sampling} & \textbf{DSC} & \textbf{SSIM} \\
\hline
Real   & ---        & --- & 0.9034 & --- \\
\hline
Missing modality & Swin-ViT  & 25  & 0.8591 & 0.7709 \\
\hline
\end{tabular}
\label{tab:modal_generation_local}
\end{table}
\begin{table}[ht]
\centering
\scriptsize
\begin{tabular*}{\textwidth}{@{\extracolsep{\fill}} llccccccc}
\hline
\textbf{Group} & \textbf{Dice\_ET} & \textbf{Dice\_TC} & \textbf{Dice\_WT} & \textbf{NSD\_ET} & \textbf{NSD\_TC} & \textbf{NSD\_WT} & \textbf{SSIM} \\
\hline
GLI & $0.82 \pm 0.21$ & $0.85 \pm 0.24$ & $0.92 \pm 0.09$ & $0.57 \pm 0.26$ & $0.51 \pm 0.27$ & $0.52 \pm 0.19$ & $0.95 \pm 0.05$ \\
MEN & $0.80 \pm 0.31$ & $0.81 \pm 0.30$ & $0.83 \pm 0.28$ & $0.59 \pm 0.32$ & $0.59 \pm 0.32$ & $0.57 \pm 0.26$ & $0.95 \pm 0.02$ \\
ALL & $0.80 \pm 0.26$ & $0.83 \pm 0.27$ & $0.88 \pm 0.20$ & $0.56 \pm 0.28$ & $0.53 \pm 0.29$ & $0.52 \pm 0.22$ & $0.95 \pm 0.04$ \\
\hline
\end{tabular*}
\caption{Performance of the missing modality generation on the test set (SOTA results). The results of the other contestants are not available, only the \href{https://www.synapse.org/Synapse:syn64153130/wiki/633062}{ranking}.}
\label{tab:modal_generation_online_test}
\end{table}

For the tumor removing we had access to an online validation platform, therefore we present them in Table \ref{tab:inpainting_online_val} the best results achieved by other contestants (which we call \textit{One-time U-Net}) and ours. The performance of the \textit{One-time U-Net} and the VS-SSIM generator are comparable. The differences in MSE, PSNR ($24.87$ vs $24.53$), and SSIM ($0.87$ vs $0.86$) are statistically marginal and fall well within one standard deviation of each other. 
The test results are presented in Table \ref{tab:inpainting_online_test}. Once again, we cannot share the results of other contestants. We achieved a new SOTA tided up with another contestant, as can be seen in the \href{https://www.synapse.org/Synapse:syn64153130/wiki/633062}{rankings} table. Visually, the proposed method produces sharper reconstructions. While the other  approach introduce oversmoothing, our model preserves high-frequency details, resulting in less blurred synthetic modalities.

\begin{table}[]
\centering
\begin{tabular*}{\linewidth}{@{\extracolsep{\fill}}llcccc}
\hline
\textbf{Name} & \textbf{Sampling} & \textbf{Network} & \textbf{MSE} & \textbf{PSNR} & \textbf{SSIM} \\
\hline
Top rank teams & 1 & U-Net based & 0.005 ±0.004     & 24.871 ±4.645     & 0.873 ±0.087 \\ 
\hline
VS-SSIM & 25  & Swin-ViT  & 0.005 ±0.004     & 24.527 ±4.718     & 0.868 ±0.089 \\    
\hline
\end{tabular*}
\caption{Performance comparison on tumor removal on the online validation set using MSE, PSNR, and SSIM metrics. The first row is the solution with best solution from another participant, and the second row our approach.}
\label{tab:inpainting_online_val}
\end{table}

\begin{table}[h!]
\centering
\begin{tabular*}{\linewidth}{@{\extracolsep{\fill}}lcccc}
\hline
\textbf{Task} & \textbf{RMSE} & \textbf{PSNR} & \textbf{SSIM} \\
\hline
Local inpainting & 0.053 ±0.027 & 26.77 ±5.21 & 0.918 ±0.089 \\
\hline
\end{tabular*}
\caption{Performance of tumor removal approach on the test set (SOTA results). Shared the same performance with other contestant. The results of the other contestants are not available, only the \href{https://www.synapse.org/Synapse:syn64153130/wiki/633062}{ranking}.}
\label{tab:inpainting_online_test}
\end{table}

\section{Conclusion}
The proposed VS-DDPM can be tailored to the available hardware, effectively alleviating the hardware constraints commonly encountered in medical institutions. This adaptability makes it an ideal solution for balancing both performance and efficiency. 
VS-DDPM achieved SOTA performance in missing modality generation and brain inpainting, as well as strong performance in MRI/CBCT to sCT translation. 
The lower performance on the MRI/CBCT to sCT translation might be related with data pre or post-processing, as well as overfitting and training instability caused by mixed-precision.
Therefore, future work should focus on increasing the stability of training, as well as focusing on developing a pipeline capable of processing the entire volume simultaneously, rather than sequentially by windows.

\begin{credits}
\subsubsection{\ackname} André Ferreira was supported by FCT - Fundação para a Ciência e Tecnologia, I.P. by project reference 2022.11928.BD and DOI identifier \url{https://doi.org/10.54499/2022.11928.BD}
. 
This work has been supported by FCT – Fundação para a Ciência e Tecnologia within the R\&D Unit Project Scope UID/00319/Centro ALGORITMI (ALGORITMI/UM), and by KITE (Plattform für KI-Translation Essen) from the REACT-EU initiative (EFRE-0801977, \url{https://kite.ikim.nrw/}).
\end{credits}

\section*{Author Contributions}
\begin{itemize}
    \setlength{\itemsep}{0pt} 
    \item \textbf{Conceptualization:} A.F., V.A., J.E.
    \item \textbf{Data curation:} A.F., N.M.
    \item \textbf{Formal analysis:} A.F., N.M., V.A., J.E.
    \item \textbf{Funding acquisition:} V.A., J.E.
    \item \textbf{Investigation:} A.F., N.M.
    \item \textbf{Methodology:} A.F., N.M., V.A.
    \item \textbf{Project administration:} A.F.
    \item \textbf{Resources:} V.A., J.E., J.K.
    \item \textbf{Software:} A.F., N.M.
    \item \textbf{Supervision:} V.A., J.E., A.F.
    \item \textbf{Validation:} A.F., N.M., B.HP.
    \item \textbf{Visualization:} A.F., N.M., B.HP.
    \item \textbf{Writing -- original draft:} A.F., N.M.
    \item \textbf{Writing -- review \& editing:} N.M., G.L., B.HP., J.K., V.A., J.E., A.F.
\end{itemize}
\bibliographystyle{splncs04}
\bibliography{bib}

\end{document}